\documentclass{article}

\PassOptionsToPackage{numbers}{natbib}
\usepackage[dblblindworkshop,final]{neurips_2025}

\workshoptitle{Machine Learning and the Physical Sciences}

\usepackage{graphicx}% Include figure files
\usepackage{svg}
\usepackage{caption}
\usepackage{subcaption}
\usepackage{amsfonts}
\usepackage{mathtools}
\usepackage{float}
\usepackage{amsmath}
\usepackage{amssymb}

\usepackage[utf8]{inputenc} % allow utf-8 input
\usepackage[T1]{fontenc}    % use 8-bit T1 fonts
\usepackage{hyperref}       % hyperlinks
\usepackage{url}            % simple URL typesetting
\usepackage{booktabs}       % professional-quality tables
\usepackage{amsfonts}       % blackboard math symbols
\usepackage{nicefrac}       % compact symbols for 1/2, etc.
\usepackage{microtype}      % microtypography
\usepackage{xcolor}         % colors

\usepackage{siunitx}

\newcommand{\todo}[1]{}
\renewcommand{\todo}[1]{{\color{red} TODO: {#1}}}

\title{Heterogeneous Point Set Transformers for Segmentation of Multiple View Particle Detectors}

\author{
  Edgar E. Robles \\
  Department of Computer Science\\
  University of California, Irvine\\
  Irvine, CA 92697 \\
  \And
  Dikshant Sagar \\
  Department of Computer Science\\
  University of California, Irvine\\
  Irvine, CA 92697 \\
  \And
  Alejandro Yankelevich \\
  Department of Physics \\
  University of California, Irvine\\
  Irvine, CA 92697 \\
  \And
  Jianming Bian \\
  Department of Physics \\
  University of California, Irvine\\
  Irvine, CA 92697 \\
  \And
  Pierre Baldi \\
  Department of Computer Science\\
  University of California, Irvine\\
  Irvine, CA 92697 \\
  \And
  For the NOvA Collaboration
}

\setcitestyle{square}

\begin{document}

\maketitle

\begin{abstract}
  %The abstract paragraph should be indented \nicefrac{1}{2}~inch (3~picas) on
  %both the left- and right-hand margins. Use 10~point type, with a vertical
  %spacing (leading) of 11~points.  The word \textbf{Abstract} must be centered,
  %bold, and in point size 12. Two line spaces precede the abstract. The abstract
  %must be limited to one paragraph.
  NOvA is a long-baseline neutrino oscillation experiment that detects neutrino particles from the NuMI beam at Fermilab. Before data from this experiment can be used in analyses, raw hits in the detector must be matched to their source particles, and the type of each particle must be identified. This task has commonly been done using a mix of traditional clustering approaches and convolutional neural networks (CNNs). Due to the construction of the detector, the data is presented as two sparse 2D images: an XZ and a YZ view of the detector, rather than a 3D representation. We propose a point set neural network that operates on the sparse matrices with an operation that mixes information from both views. Our model uses less than 10\% of the memory required using previous methods while achieving a 96.8\% AUC score, a higher score than obtained when both views are processed independently (85.4\%).
\end{abstract}

\section{Introduction}

Experiments in the physical sciences often create large amounts of data, which is hard to process at scale by human experts. The advent of high quality machine learning models has improved performance across many of these data processing tasks~\cite{baldi2021science}, but with an increase in quality, there has also been an increase in computational costs. Even large experimental collaborations in the field of particle physics often face strict limits in resource utilization.

\begin{figure}
    \hfill
    \begin{subfigure}[t]{0.4\textwidth}
        \centering
        \includegraphics[width=1.05\textwidth]{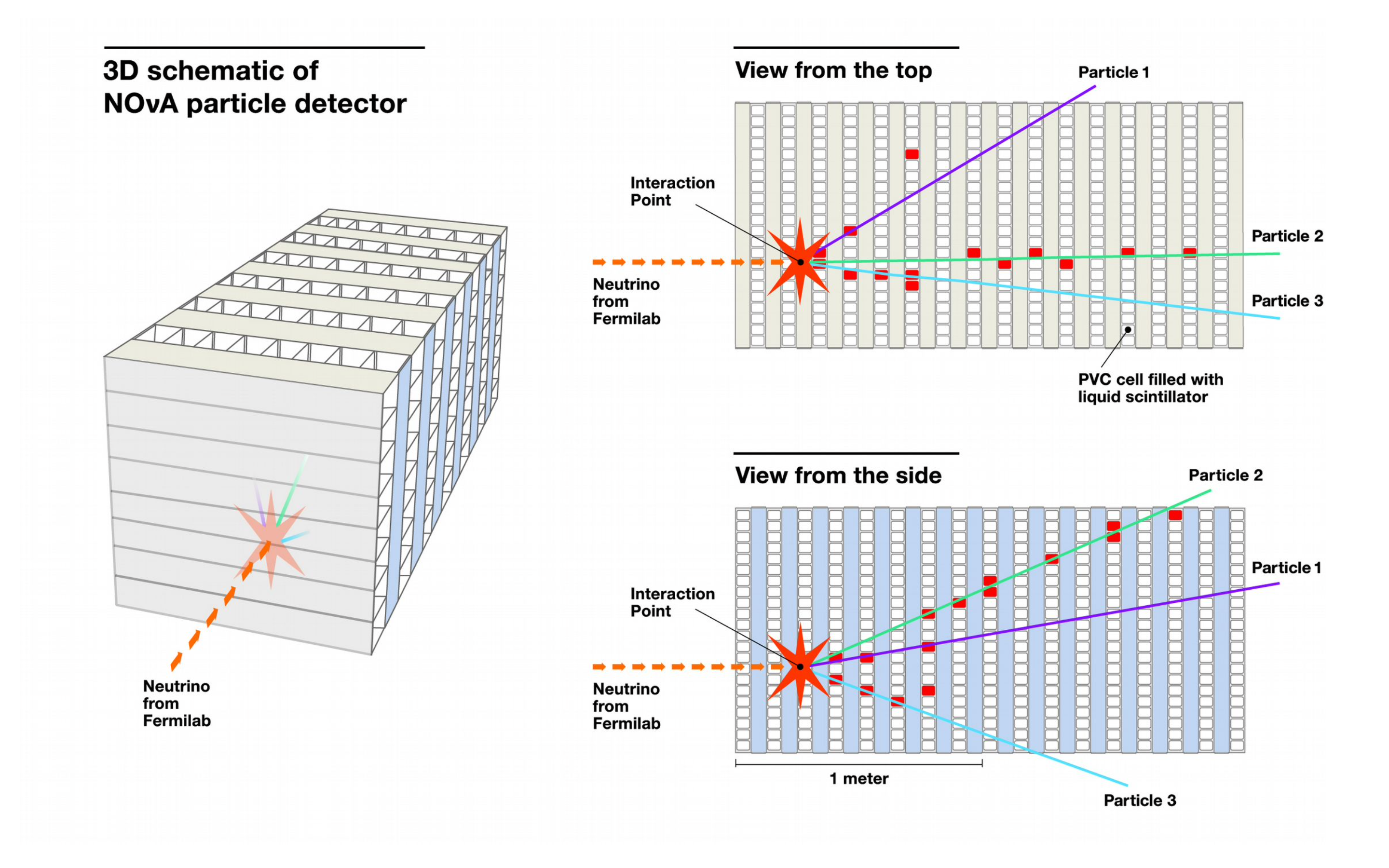}
        \caption{Schematic of NOvA detector and generation of top and side views from vertical and horizontal planes respectively.}
        \label{fig:schematic}
    \end{subfigure}
    \hfill
    \begin{subfigure}[t]{0.55\textwidth}
        \centering
        \includegraphics[width=0.75\textwidth]{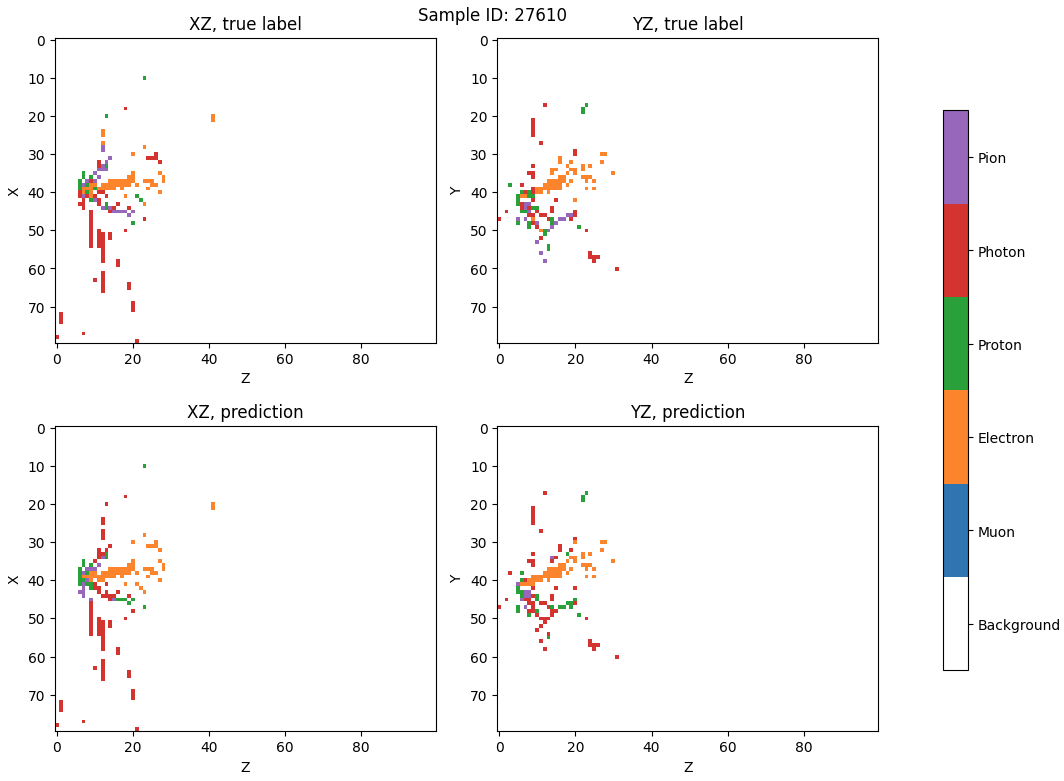}
        \caption{Example event display image of a neutrino interaction producing a prominent electron shower and secondary particles shown with true particle classes (top) and HPST predicted classes (bottom).}
        \label{fig:interactions}
    \end{subfigure}
    \hfill
    \caption{Visual representations of the NOvA detector and data.}
\end{figure}

NOvA \cite{nova_technical} is a long-baseline neutrino oscillation experiment in which neutrinos from Fermilab's NuMI beam interact in both the near detector (ND) at in Batavia, IL and larger far detector (FD) in Ash River, MN spaced \SI{810}{km} apart. The detectors consist of long $\SI{4}{cm} \times \SI{6}{cm}$ polyvinyl chloride (PVC) plastic extrusions, referred to as "cells", filled with liquid scintillator. These cells are arranged in planes such that the cells' orientations alternate horizontally and vertically between neighboring planes (Figure \ref{fig:schematic}). The full FD considered here is $\SI{16}{m} \times \SI{16}{m} \times \SI{60}{m}$ with 344064 cells. When neutrinos interact within the detector, the resulting particles, e.g. pions, photons, protons, electrons and muons, produce scintillation light as they travel through the detector and appear as tracks or showers referred to as "prongs" (Figure \ref{fig:interactions}) \cite{simulation}. The task at hand is then to perform instance segmentation over these prongs as well as to classify each detection into its corresponding particle type.

The generated images are often very sparse, consisting of an empty background in most of the image except for a few prongs. When performing computations such as the ones used in segmentation machine learning models, sparse matrices have to be converted into dense matrices, which can slow down training and inference. There have been implementations of differentiable convolution operations on sparse matrices, such as Nvidia's MinkowskiEngine~\cite{choy20194d}. However, the operations need to approximate a convolution in order to save memory. An alternative to using sparse matrices is representing the sparse image as a point cloud, which allows the coordinates and values to be processed.

The design of the NOvA detectors only allows images to be produced over two planes: the XZ plane (top view) and YZ plane (side view), in which each view only consists of vertical or horizontal cells, respectively (Figure~\ref{fig:schematic}). Therefore, the Y position of a hit in a vertical plane fundamentally cannot be known. The existing method for the instance segmentation task approaches this by applying the fuzzy k-means clustering algorithm on each view independently followed by matching instances across the two views \cite{reco}. Similarly for the particle classification task,
existing methods use two independent CNNs to embed each image~\cite{eventcvn, prongcvn, regcnn}
or use each view as a channel to an image~\cite{transformercvn}. We can extend the framework of point transformers by using heterogeneous attention~\cite{hu2020heterogeneous}.
In this paper we leverage point set transformers, along with implementing a novel method to allow information to flow between both views of the dataset, allowing for a model that is able to draw information from both views. The model proposed is known as a heterogeneous point set transformer (HPST).

\section{Methods}
\subsection{Notation}
Consider a dataset $\mathcal{X}$ of size $N$, where each sample $X^{(i)}$ represents an event from the particle detector. Each event $X^{(i)}$ is split into $M$ views, each view denoted by $X^{(i,j)}$ In the case of the NOvA detector, we have $M=2$ views. Each view has a variable number of detections $K^{(i,j)}$. Each detection is described by coordinates $x^{(i,j)}_k \in \mathbb{R}^c$ and values $v^{(i,j)}_k \in \mathbb{R}^d$. 

For each pair of points we define an intra-view distance $d_{jj}(x^{(i,j)}_k,x^{(i,j)}_{k'})$ for points within the same view and therefore vector space and an inter-view distance $d_{jj'}(x^{(i,j)}_{k},x^{(i,j')}_{k'})$ for points between different views. Additionally, based on these distances we will define an edge $e^{(i)}_{k,k'} \in \{0, 1\}$ which connects two nodes that may be in the same or different views.

\subsection{Heterogeneous attention}
In order to calculate attention, we need to encode each point into a set of keys, queries and values, each of these is done by taking the input point $x^{(i,j)}_k$ and multiplying it by a linear layer as such: $Q^{(i,j'\to j)}_k = W^{(j'\to j)}_Q x^{(i,j)}_k$, $K^{(i,j'\to j)}_k = W^{(j'\to j)}_K x^{(i,j)}_k$, $V^{(i,j'\to j)}_k = W^{(j'\to j)}_V x^{(i,j)}_k$.

We use the inter-view distance to build a nearest neighbors graph, then, for each point we calculate the query $Q^{(i,j'\to j)}_k$, i.e., the query on point $k$ from view $j'$ to view $j$ on sample $i$, and then for each of its neighbors $k'$ we calculate both $K^{(i,j'\to j)}_{k'}$ and $V^{(i,j'\to j)}_{k'}$, that is, the key and values on point $k'$ from view $j'$ to view $j$ on sample $i$. Using these, we  can calculate
$
w^{(i,j'\to j)}_{kk'} = Q_{k}^{(i,j'\to j)T}K^{(i,j'\to j)}_{k'}.
$
RPE (relative positional encoding, $\mathrm{RPE}(x,y) = W(x-y)$, where $W$ is a learnable linear layer as used in~\cite{wu2022}) is not used between different views. This weight is then normalized using a softmax operation over its neighbors and then used in a weighted sum to calculate the output of the attention module,
$$
{h'}^{(i,j)}_{k} = \sum_{k'} \mathrm{softmax}_{\ell}(w^{(i,j'\to j)}_{k\ell})_{k'} V^{(i,j'\to j)}_{k'}.
$$

Pooling is done within the same view, using a voxel pooling method~\cite{simonovsky2017}, creating a grid and then averaging out the values of all the points within each point of the grid, and positioning the point in the barycenter of all the points within the created voxel. % cite pytorch geometric (https://arxiv.org/abs/1704.02901)
Unpooling is performed using skip connections, the points are upsampled to the same coordinates that they were previously pooled from, only using information from the same view. 

\subsection{Architecture}
\begin{figure}
    \centering
    \includegraphics[width=0.7\textwidth]{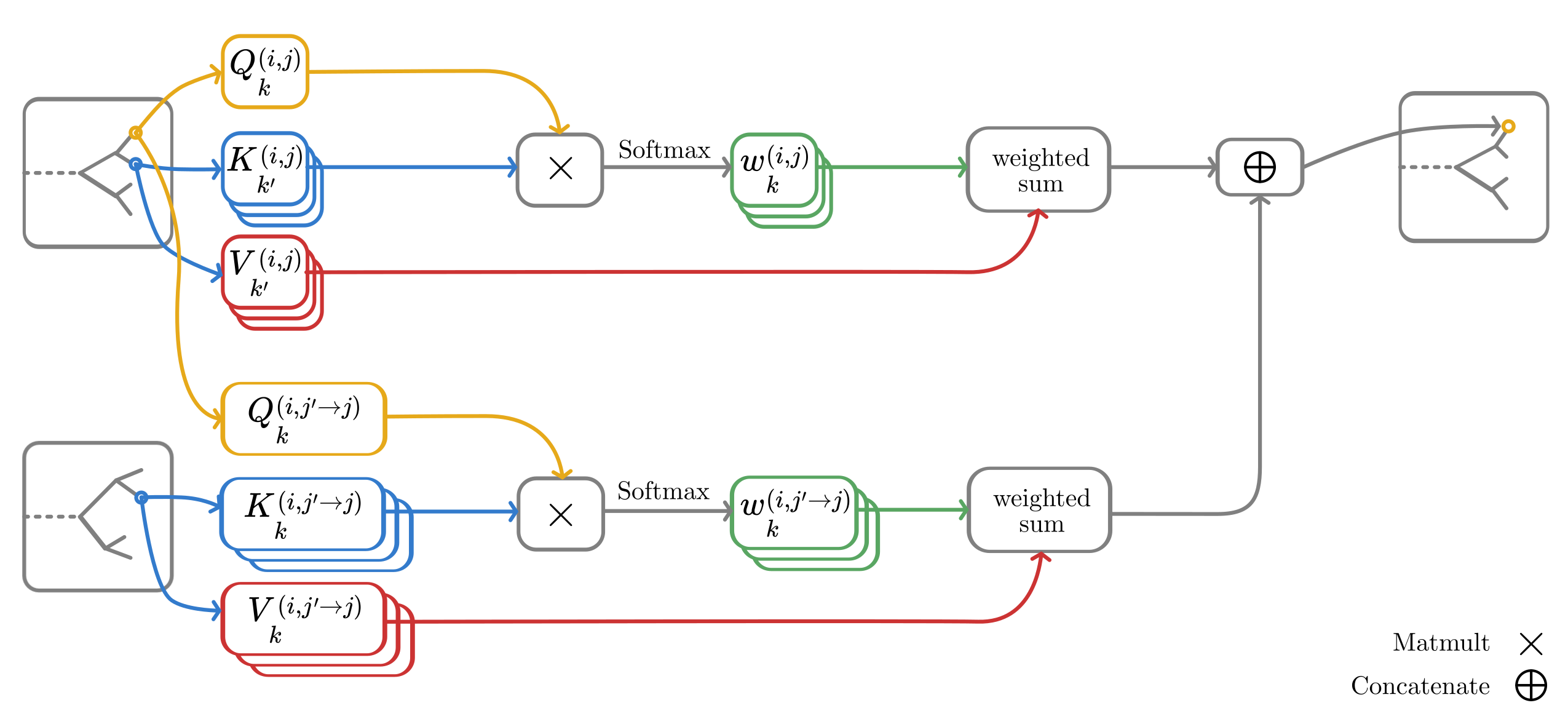}
    \caption{Block diagram of the attention mechanism. The top path describes the intra-view attention mechanism, and the bottom path describes the inter-view mechanism.}\label{fig:attn}
\end{figure}

The network is structured like a UNet. The UNet is divided into $2n$ stages. Each stage contains $m$ blocks, where each block has an attention module. Following each stage in the first $n$ stages is a pooling step, which reduces the number of points. The next $n$ stages are followed by an unpooling stage, which uses the point coordinates from a past block, as well as concatenates the features from the past block. The dimensionality of the embeddings is doubled at each stage during the first half and halved at each stage during the second half. Intra-view attention is calculated on each stage in order to ensure that the information mixing between views is done locally (in the earlier stages) as well as globally (in the later stages). 

\subsection{Loss function}
The network performs two tasks simultaneously: an instance segmentation, selecting separate prongs from each other; and a semantic segmentation, classifying each detection into a particle type. As such, the loss function used is separated into two parts: $\mathcal{L} = \lambda\mathcal{L}_{\mathrm{sem}} + (1-\lambda)\mathcal{L}_{\mathrm{ins}}$.
Semantic segmentation is a simple classification problem, so we use multi-class cross-entropy to calculate this loss:
$
\mathcal{L}_{\mathrm{sem}} = \sum_{X^{(i)} \in \mathcal{X}} \sum_{X^{(i)} \in x^{(i,j)}_{k}} \mathrm{CE}\left(\mathrm{softmax}_{k'}\left(f(X_i)^{(i,j)}_{k'}\right),y^{(i,j)}_{k}\right),
$
where $y^{(i,j)}_{k}$ is the correct semantic label of the detection. 

Instance segmentation is done by minimizing the loss calculated by the best assignment between the predicted labels and the real labels. If point $x^{(i,j)}_{k}$ belongs to the segment $L^{(i,j)}_{k}$, then the loss calculated is
$
\mathcal{L}_{\mathrm{ins}} = \sum_{X^{(i)} \in \mathcal{X}} \min_{\phi \in \Sigma} \sum_{x^{(i,j)}_{k} \in X^{(i)}} \mathrm{CE}\left(\mathrm{softmax}_{k'}\left(f(X_i)^{(i,j)}_{k'}\right),\phi\left(L^{(i,j)}_{k}\right)\right),
$
where $\Sigma$ is the set of all permutations of labels, allowing a unique assignment of one label to another. The optimal assignment of the labels is solved using a linear sum assignment solver.

\section{Experiments and Results}
\subsection{Experiments}
The dataset used consists of neutrino interaction simulations generated by the NOvA collaboration for the purpose of preparing event reconstruction methods for future large-scale productions. This dataset contains 9246712 events, each event containing 70 hits on average in $ 2 \times 80 \times 100$ images. 

%\subsection{Experiments}
Three models were trained and evaluated: a Mask-RCNN~\cite{he2017mask} based model, GAT \cite{velickovic2017graph}, and our heterogeneous point set transformer. 
We performed a hyperparameter search: using the number of neighbor connections (4, 8), the number of stages of the neural network (2, 3, 4), the size of the embeddings inside the neural network (128, 256, 512), and the learning rate (between 1e-4 and 1e-1). The training and testing was done on a server using an Intel(R) Xeon(R) CPU E5-2640 v4 @ 2.40GHz, 503G of RAM, and 4xNVIDIA Titan V.  The hyperparameter sweep was peformed over 8 epochs, using 1\% of the dataset. The model with the best accuracy on the segmentation's class labels was selected as the one with the best hyperparameters. The resulting models with the best hyperparameters were trained for 24 epochs. The code for these experiments can be found at \url{https://github.com/erobl/hpst}.

\subsubsection{Classification and Segmentation accuracy}

\begin{figure}
    \centering
    \includegraphics[width=0.9\textwidth]{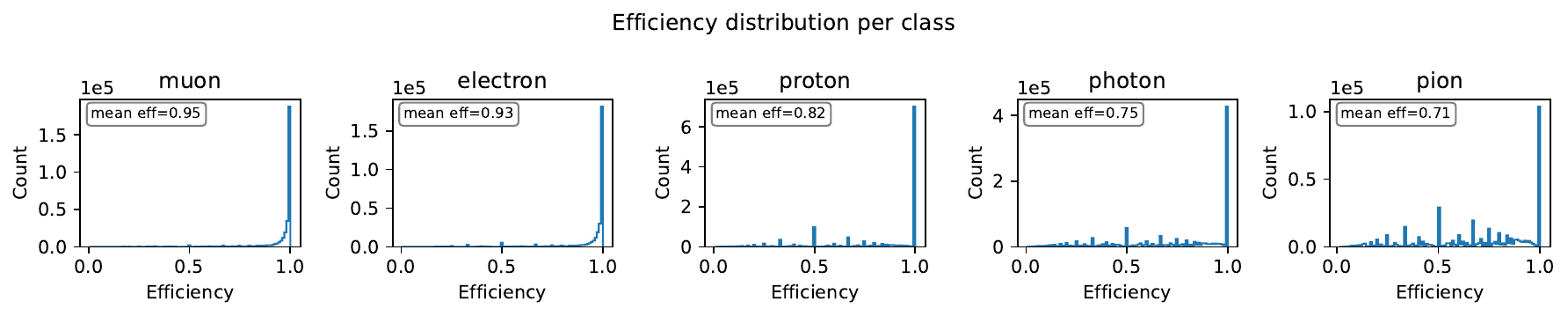}\vspace*{-4pt}
    \includegraphics[width=0.9\textwidth]{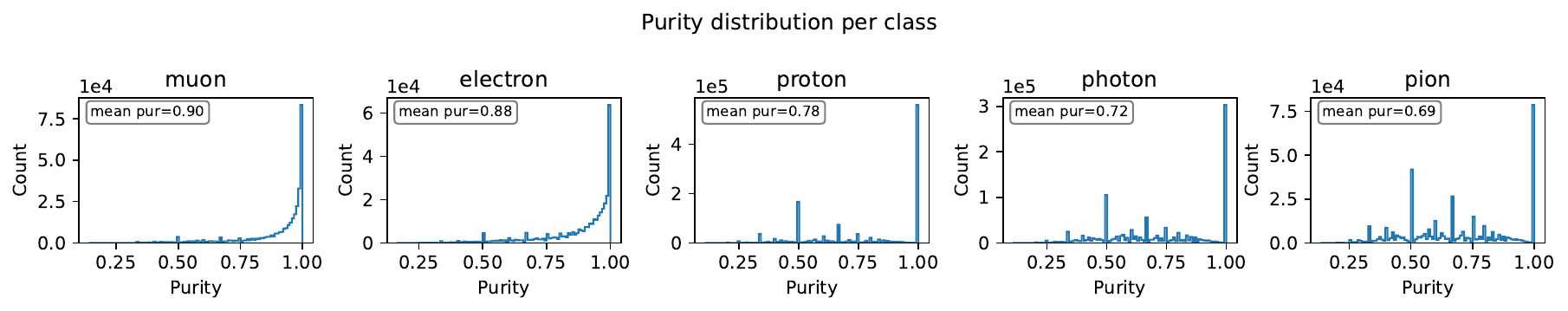}
    \caption{Distribution of prong efficiency and purity}\label{fig:dist}
\end{figure}
We present the evaluation of each model in Table~\ref{tbl:speedmemory}.
For each prong, we calculated the efficiency (recall) and purity (precision) of the classification, allowing for multiple predicted prongs to be assigned to a single prong. 
Figure~\ref{fig:dist} shows the distribution of purity and efficiency for each class. As we can see, the segmentation results are generally good, especially in the majority classes (muons and electrons).
Figure~\ref{fig:interactions} shows a qualitative example from the test set. This sample presents a mostly correct classification, with some confusion for secondary particles producing very few hits.

\subsubsection{Speed and memory usage}
\begin{table}
  \caption{Speed and memory usage for each model compared to their performance.}
  \label{tbl:speedmemory}
  \centering
  \begin{tabular}{lllll}
    \toprule
    Model     & Memory & Time per   & OVR AUC & Segmentation\\
    & usage (MiB) & sample (s) & & accuracy\\
    \midrule
    R-CNN & $151.9 \pm 11.950$ & $3.1449 \pm 0.0660$ & 0.732 & 0.343 \\
    GAT & $29.10 \pm 0.001$ & $0.3911 \pm 0.0010$    & 0.854 & 0.659 \\
    HPST (ours)   & $29.35 \pm 0.559$ & $0.4636 \pm 0.0032$    & 0.968 & 0.835 \\
    \bottomrule
  \end{tabular}
\end{table}
We benchmarked the three models by running inference on 100 samples, with a batch size of 1, measuring the peak memory increase between the start of inference and the end of inference, to remove as much overhead as possible. We evaluated the time it takes for these 100 inferences and the memory used in each of them. 
Memory usage is greatly reduced compared to the regular CNN model. Our model is able to obtain a superior performance over the other models while still keeping a comparable memory usage.

\section{Limitations and Conclusions}
While the claims of memory efficiency will generally hold true, for different datasets this might not be the case. The representation of a sparse matrix is more efficient than a dense matrix until a certain point, where the storage of the coordinates becomes bigger than just storing a dense matrix. Point set operations can also greatly increase in complexity as the number of points grows, resulting in a much slower algorithm. However, these are not the regimes found in the the NOvA experiment. 
HPSTs strike a balance between memory usage, time, and performance that is fitting for its application, they are able to train a stronger model with the same amount of parameters due to sharing information between the views, which in turn boosts the performance of the model.

\begin{ack}
This document was prepared by the NOvA collaboration using the resources of the Fermi National Accelerator Laboratory (Fermilab), a U.S. Department of Energy, Office of Science, Office of High Energy Physics HEP User Facility. Fermilab is managed by FermiForward Discovery Group, LLC, acting under Contract No. 89243024CSC000002.
\end{ack}

\bibliography{bib}

\end{document}